\documentclass{article} 
\usepackage{iclr2027_conference,times}

\usepackage{amsmath,amsfonts,bm}

\def\eqref#1{equation~\ref{#1}}

\def\1{\bm{1}}

\DeclareMathAlphabet{\mathsfit}{\encodingdefault}{\sfdefault}{m}{sl}
\SetMathAlphabet{\mathsfit}{bold}{\encodingdefault}{\sfdefault}{bx}{n}

\usepackage{hyperref}
\usepackage{url}

\usepackage{graphicx}
\usepackage{float}
\usepackage{placeins}
\usepackage{booktabs}
\usepackage[table]{xcolor}
\usepackage{tabularx}

\usepackage{algorithm}
\usepackage{algpseudocode}

\usepackage{kotex}

\AddToHook{env/figure/begin}{\setlength{\parskip}{0pt}}
\AddToHook{env/table/begin}{\setlength{\parskip}{0pt}}

\providecommand{\method}{TaSQ}

\title{Tailoring the Quantization Space for 1-Bit KV Cache Compression}

\author{%
  Minsoo Cheong\thanks{Equal contribution.} \\
  Seoul National University\\
  \texttt{icycle0409@snu.ac.kr} \\
  \And
  Donghyun Son\footnotemark[1] \\
  Stanford University\\
  \texttt{dhson@stanford.edu} \\
  \And
  Sungjoo Yoo\thanks{Corresponding author.} \\
  Seoul National University\\
  \texttt{sungjoo.yoo@gmail.com} \\
}

\usepackage{hyperref}
\usepackage{fontawesome5}

\iclrfinalcopy
\begin{document}

\maketitle

\begin{abstract}

The key-value (KV) cache becomes a major memory bottleneck in long-context LLM inference, placing substantial pressure on memory capacity and bandwidth.
To mitigate this bottleneck, vector quantization (VQ) has emerged as a promising approach for aggressive KV cache compression.
However, existing VQ methods degrade substantially in the 1-bit regime.
At such extreme compression, each codebook must represent a larger group of channels with a limited set of centroids, making effective use of its capacity increasingly challenging.
To address this, we introduce \textbf{\method{}}, which tailors the VQ target space by combining query-guided channel weighting, cross-head normalization, and covariance-aware channel grouping to better reflect the error sensitivity and statistical structure of cached activations.
Since these transforms are RoPE-compatible and can be easily merged into projection weights and codebooks, \method{} preserves the conventional VQ lookup structure and adds negligible serving overhead.
Across general, long-chain-of-thought reasoning, and long-context retrieval benchmarks, \method{} consistently outperforms existing low-bit KV cache VQ baselines while preserving reasoning stability.
On a single RTX 6000 Ada GPU, its SGLang implementation supports up to $14\times$ larger batch sizes and achieves $1.87\times$ higher peak throughput compared to the BF16 baseline.
\end{abstract}
{\centering
\href{https://github.com/mscheong01/tasq}{
  \textcolor{blue}{
    \faGithub\ 
    \texttt{\textbf{Code}: https://github.com/mscheong01/tasq}
  }
}
\par}
\section{Introduction}

Modern LLM applications increasingly operate over interaction histories and large inputs,
making long-context inference an important serving workload.  Supporting such contexts requires
storing the key and value activations of preceding tokens in the key--value (KV) cache.  Because
the cache grows linearly with sequence length and must be repeatedly accessed during autoregressive
decoding, KV cache capacity and memory bandwidth become central bottlenecks in long-context
serving.

KV cache compression directly targets these bottlenecks through various approaches, including
token eviction \citep{zhang2023h2o,li2024snapkv}, cache merging \citep{wang2024model}, and
quantization \citep{liu2024kivi,hooper2024kvquant}.  Among them, quantization is one of the most
widely studied and adopted approaches.
In particular, vector quantization (VQ) is well suited to ultra-low-bit settings, as it jointly represents multiple channels with learned codewords that capture their multidimensional structure.
Existing approaches learn codebooks over contiguous channel
groups \citep{zhang2024kv} or normalize activations to use a calibration-free global codebook
\citep{son2026nsnquant}.  Yet preserving model quality as the rate approaches 1 bit per channel
remains challenging.

To achieve high quantization quality at these rates, each codebook must effectively represent larger groups of channels with a limited set of centroids.
We tackle this challenge by accounting for how quantization errors affect attention logits and how inter-channel dependencies can be exploited by VQ, based on two key insights.
First, reconstruction errors in different channels have unequal effects on $QK^\top$.
Second, VQ can exploit inter-channel dependencies only when the corresponding channels are represented by the same codebook.

Motivated by this, we introduce TaSQ (\textbf{Ta}ilored \textbf{S}pace Vector \textbf{Q}uantization), an LLM KV Cache VQ method that tailors the quantization space for accurate ultra-low-bit compression.
\method{} uses query-guided weighting to emphasize sensitive channels and covariance-aware grouping to assign dependent channels to the same local VQ group.
A compact cross-head normalization further handles token-level magnitude outliers.  
The method employs only channel-wise scaling and permutation, without a dense
rotation.
These operations are mostly absorbed into the weight projection and codebooks, while runtime reconstruction operations are fused into the serving kernels.
\method{} therefore retains the conventional VQ lookup structure and incurs negligible runtime overhead.

Across general, reasoning, and long-context retrieval benchmarks, \method{} substantially
improves accuracy over existing VQ baselines in the 1-bit regime.  In our SGLang serving
implementation, \method{} reduces the KV cache footprint and bandwidth demand, expanding the
available KV cache pool by \(12.48\times\) to support larger batches and enabling up to \(1.87\times\) higher throughput
compared to the full-precision BF16 baseline.
These results demonstrate that tailoring the VQ target space is an effective approach to accurate and efficient ultra-low-bit KV cache compression.


\section{Preliminaries}

\subsection{LLM Generation and KV Cache}
Most modern large language models (LLMs) are based on the decoder-only Transformer
architecture~\citep{vaswani2017attention} and generate text autoregressively.
Under causal masking, the keys and values of preceding tokens remain unchanged as
new tokens are generated, allowing them to be stored in a key--value (KV) cache
and reused across decoding steps.
However, the cache grows linearly with sequence length and must be read at every
decoding step.
As context length increases, storing and accessing the KV cache places growing
pressure on memory capacity and bandwidth.

\subsection{Vector Quantization}

Vector quantization (VQ) compresses a multi-dimensional vector by replacing it
with the index of a codeword from a finite codebook. Given a codebook
$\mathcal{C}=\{c_1,\ldots,c_{K}\}$ and an input $x\in\mathbb{R}^{g}$, VQ compresses the given input by assigning the nearest codeword:
\[
    z(x)=\arg\min_j \|x-c_j\|_2^2,
    \qquad
    \hat{x}=c_{z(x)}.
\]
The codebook is typically learned by applying k-means to representative samples, with the resulting cluster centroids serving as codewords.

When a $g$-dimensional vector is represented by one of $K$ codewords, the index
cost is $\log_2 K / g$ bits per channel. 
Since VQ represents multiple channels jointly
and can exploit structure in their joint distribution, it is 
particularly attractive for ultra-low-bit compression.

\begin{figure}[!t]
    \setlength{\abovecaptionskip}{3pt}
    \centering
    \includegraphics[width=\linewidth]{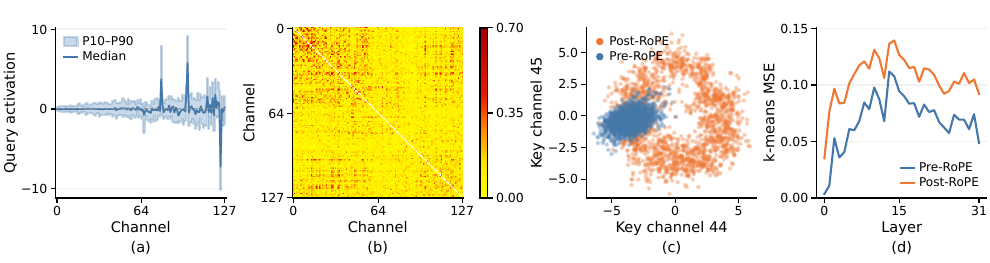}
    \vspace{-0.5em}
    \caption{(a) Per-channel query medians and P10--P90 intervals.
    (b) Absolute inter-channel key correlations, with self-correlations masked.
    (c) Key distributions before and after RoPE for channels 44/45, using identical sampled tokens.
    (d) K-means reconstruction MSE on the fitting samples across all 32 layers,
    with matched group layouts and codebook sizes.
    All panels use a single 2,048-token GPQA window from Llama-3.1-8B;
    (a--c) show layer 15, KV head 0.}
    \label{fig:motivation-overview}
\end{figure}

\section{Motivation}
\label{sec:method:motivation}

Prior works have observed pronounced structure along the channel dimension of keys,
including outlier channels and inter-channel dependencies
\citep{liu2024kivi, hooper2024kvquant, zhang2024kv, xu2025think}.
In this work, we likewise focus on the behavior of \textbf{key} channels.
In particular, we observe that key channels differ in their sensitivity to queries, exhibit non-uniform correlations with one another, and undergo substantial distributional changes after RoPE.
These observations motivate us to \textbf{tailor the quantization space to better reflect the structure of key channels}.
In the matched analysis in Appendix~\ref{app:key-value-structure}, values, in contrast, exhibit more uniform channel-wise loss sensitivity within each head and weaker average inter-channel correlations than keys. We therefore focus on keys and leave the design of a VQ space tailored to values to future work.

\subsection{Sensitivity of Key Channels to Queries}
\label{sec:motivation-channel-importance}

Quantization errors in different key channels do not affect attention scores equally.
In the query--key dot product, an error in each key channel is scaled by the corresponding query activation.
As shown in Figure~\ref{fig:motivation-overview}a, query activation distributions vary substantially across channels, with several channels exhibiting particularly large activation ranges.
These differences imply that key channels have different sensitivities to quantization error, motivating the query-guided channel weighting introduced in Section~\ref{sec:query-guided-weighting}.

\begin{samepage}
\subsection{Inter-channel Correlation}
\label{sec:motivation-channel-correlations}

As shown in Figure~\ref{fig:motivation-overview}b, key channels exhibit varying dependencies with one another.
While some channel pairs are strongly correlated, others are nearly independent.
Since VQ represents multiple channels jointly with a shared codebook, its ability to exploit these dependencies depends on which channels are grouped together.
This motivates covariance-aware channel grouping, which we introduce in Section~\ref{sec:channel-grouping}.

\end{samepage}

\subsection{Pre-RoPE vs.\ Post-RoPE Keys}
\label{sec:motivation-pre-post-rope}

VQ is effective when the input distribution can be accurately represented by a finite set of centroids.
However, RoPE rotates keys by position-dependent angles, substantially spreading their distribution.
As shown in Figure~\ref{fig:motivation-overview}c, the relatively compact pre-RoPE key distribution becomes widely dispersed after RoPE, making it more difficult for a shared codebook to cover the space efficiently.
Figure~\ref{fig:motivation-overview}d confirms this effect: pre-RoPE VQ achieves lower reconstruction error across all 32 layers, with a 35\% lower total reconstruction error.
Consequently, we design \method{} to apply VQ to pre-RoPE keys.


\FloatBarrier

\begin{figure}[t!]
    \centering
    \includegraphics[width=0.95\linewidth]{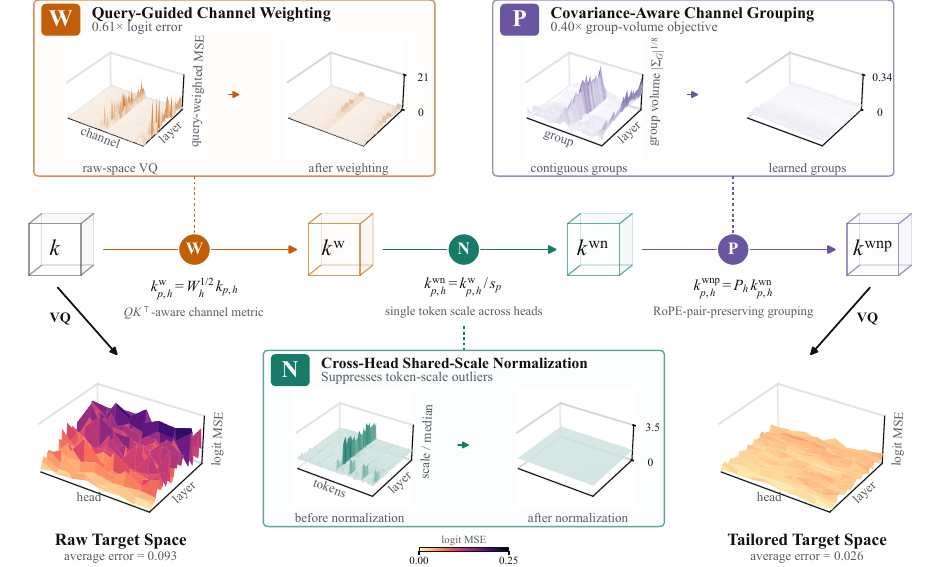}
    \caption{Overview of \method{}.  It tailors the VQ target space through query-guided channel
weighting, cross-head shared-scale normalization, and covariance-aware grouping that preserves
RoPE pairs, thereby reducing quantization error.}
    \label{fig:method-overview}
\end{figure}

\section{Method}
\label{sec:method}

\paragraph{Overview.}
An overview of \method{} is illustrated in Figure~\ref{fig:method-overview}.
During calibration, we apply query-guided channel weighting, normalization,
and covariance-aware channel grouping to pre-RoPE keys, then fit a VQ codebook
for each group in the resulting space. At inference time, keys are encoded as codebook indices
and a scale, then reconstructed for attention with RoPE applied on the fly.

\paragraph{Notation.}

We use $H$ for the number of KV heads and $d$ for the head dimension.
For KV head $h$, $k_{p,h}$ denotes the pre-RoPE key at position $p$,
$R_p$ its RoPE rotation matrix, and $\bar q_{t,h}$ a post-RoPE query at position $t$.
Superscripts $\mathrm{w}$, $\mathrm{n}$, and $\mathrm{p}$ denote weighting, normalization, and permutation, respectively, and accumulate from left to right: $k^{\mathrm{wnp}}$ is the weighted, normalized, and permuted key.
We denote the calibration corpus by $\mathcal D_{\mathrm{cal}}$ and use it as a superscript on expectations and covariances to indicate empirical estimation over this corpus.
We omit the layer index since the same procedure is applied to all attention layers.

\subsection{Query-Guided Channel Weighting}
\label{sec:query-guided-weighting}

Motivated by the channel-wise variation in query distributions observed in
Section~\ref{sec:motivation-channel-importance}, we derive channel weights from the
query--key dot-product error and use them to transform keys into a weighted space.

Let $\delta_{p,h} = k_{p,h} - \hat{k}_{p,h}$ denote the pre-RoPE key quantization error at position $p$ for KV head $h$. For a post-RoPE query $\bar{q}_{t,h}$ at position $t$, the squared dot-product error with the post-RoPE key is
\begin{equation*}
    \left(\bar{q}_{t,h}^{\top} R_p \delta_{p,h}\right)^2
    =
    \delta_{p,h}^{\top} A_{t,p,h} \delta_{p,h},
    \qquad
    A_{t,p,h}
    =
    R_p^{\top} \bar{q}_{t,h} \bar{q}_{t,h}^{\top} R_p.
\end{equation*}

To obtain a position- and token-independent summary of query–key sensitivity, we approximate $A_{t,p,h}$ by its average over valid causal query–key pairs in the calibration data:
\begin{equation*}
    \tilde{A}_h
    =
    \mathbb{E}_{t,p}^{\mathcal D_{\mathrm{cal}}}\left[A_{t,p,h}\right].
\end{equation*}

We then approximate $\tilde{A}_h$ with its diagonal to construct a channel-wise transform that preserves the native RoPE-pair structure:
\begin{equation*}
    W_h = \operatorname{diag}(\tilde{A}_h).
\end{equation*}
Appendix~\ref{app:diag-w} provides further analysis of this approximation. This yields a surrogate for the expected squared dot-product error:
\begin{equation}
\begin{aligned}
\mathbb{E}_{t,p}\!\left[
\delta_{p,h}^{\top}A_{t,p,h}\delta_{p,h}
\right]
&\approx
\mathbb{E}_{t,p}\!\left[
\delta_{p,h}^{\top}W_h\delta_{p,h}
\right]
=
\mathbb{E}_{t,p}\!\left[
\left\|W_h^{1/2}\delta_{p,h}\right\|_2^2
\right].
\end{aligned}
\label{eq:weighted-error-surrogate}
\end{equation}

We therefore transform each pre-RoPE key into the weighted space:
\begin{equation*}
    k^{\mathrm{w}}_{p,h} = W_h^{1/2} k_{p,h}.
\end{equation*}

\begin{figure}[!t]
    \centering
    \includegraphics[width=\linewidth]{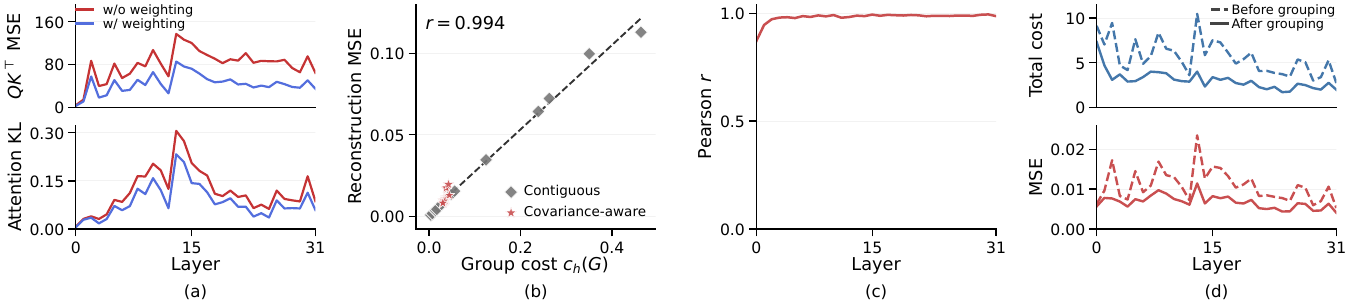}
    \setlength{\abovecaptionskip}{3pt}
    \caption{
    (a) $QK^\top$ MSE (top) and attention KL divergence (bottom) with and without
    query-guided weighting.
    (b) Relationship between the group cost $c_h(G)$ and VQ reconstruction MSE
    at layer 15, KV head 0.
    (c) Pearson correlation between group cost and reconstruction MSE across layers,
    averaged over 8 KV heads.
    (d) Total group cost (top) and VQ reconstruction MSE (bottom) under contiguous
    and covariance-aware grouping.
    Results are obtained with Llama-3.1-8B using disjoint GPQA samples for
    calibration and evaluation.
    }
    \label{fig:method-validation}
\end{figure}

Since the expected squared reconstruction error in the weighted space is equal to the surrogate in Eq.~\ref{eq:weighted-error-surrogate},
Euclidean VQ in this space incorporates query sensitivity directly into its reconstruction objective.
Consistent with this interpretation, Figure~\ref{fig:method-validation}(a) shows that query-guided weighting consistently lowers $QK^\top$ error and attention KL divergence across layers.

\subsection{Cross-head shared-scale normalization}
\label{sec:method:normalization}

Following \citet{son2026nsnquant}, we apply token-wise normalization to suppress outlier tokens before VQ.
While the prior work's head-wise normalization stores a separate scale for each token and KV head, we find that using a single RMS scale across all $H$ KV heads is sufficient.
For the
weighted keys $k^{\mathrm{w}}_{p,h}$, we compute
\begin{equation*}
  s_p=
  \left(
    \frac{1}{Hd}\sum_{h=1}^{H}\lVert k^{\mathrm{w}}_{p,h}\rVert_2^2
  \right)^{1/2},
  \qquad
  k^{\mathrm{wn}}_{p,h}=k^{\mathrm{w}}_{p,h}/s_p.
\end{equation*}
With $b_s$-bit scales, NSNQuant's
head-wise scaling costs $b_s/d$ bits per channel, whereas ours costs $b_s/(Hd)$, a factor-$H$
reduction.  We store $s_p$ in FP16 ($b_s=16$), retaining high scale resolution at a small rate overhead.
Appendix~\ref{app:shared-scale} provides further analysis.

\subsection{Covariance-Aware Channel Grouping}
\label{sec:channel-grouping}


Motivated by the inter-channel dependencies observed in
Section~\ref{sec:motivation-channel-correlations}, we seek to group channels
so that each VQ codebook can better exploit their joint structure.
We therefore introduce a covariance-aware grouping criterion for the weighted
and normalized key space.

For each KV head $h$, we estimate the covariance of the weighted and normalized keys from the calibration dataset:
\begin{equation*}
    \Sigma_h^{\mathrm{wn}}
    =
    \operatorname{Cov}^{\mathcal D_{\mathrm{cal}}}\left(k_h^{\mathrm{wn}}\right)
\end{equation*}
For a candidate channel group $G$, we define the following cost as a proxy for quantization error:
\begin{equation*}
    c_h(G)
    =
    \det\left(
        \Sigma_h^{\mathrm{wn}}[G,G] + \varepsilon I
    \right)^{1/|G|},
\end{equation*}
where $\Sigma_h^{\mathrm{wn}}[G,G]$ is the principal submatrix indexed by $G$, and $\varepsilon > 0$ is added for numerical stability.

This is motivated by the rate--distortion behavior of VQ. Under a Gaussian model, high-rate quantization theory gives the asymptotic scaling
$D_G \propto K^{-2/|G|}\det(\Sigma_h^{\mathrm{wn}}[G,G])^{1/|G|}$
for the optimal mean squared reconstruction error of a $K$-centroid codebook~\citep{gersho2012vector}.
Since $|G|$ and $K$ are fixed across groups, the only group-dependent term is the
covariance determinant, which motivates $c_h(G)$ as a proxy for quantization error.


Based on this cost function, we formulate a grouping problem as an optimization problem.
Let $g$ denote the number of channels quantized by each vector codebook. We seek a partition $\mathcal{G}_h$ of the $d$ channels that minimizes
\begin{equation*}
    \min_{\mathcal{G}_h}
    \sum_{G \in \mathcal{G}_h} c_h(G),
    \qquad
    |G| = g,
\end{equation*}
subject to each RoPE pair belonging to the same group for efficient online RoPE application.

We approximately solve this grouping problem using \textbf{hierarchical matching}. Starting from individual RoPE pairs, we assign each pair of current groups a merge cost given by $c_h(G_a \cup G_b)$ and find a minimum-weight perfect matching. We merge the matched groups and repeat until each group contains $g$ channels. Further details are provided in Appendix~\ref{app:partition-algorithm}.

The resulting partition defines a permutation matrix $P_h$ that places the channels of each group in a contiguous block. We apply this permutation to the weighted and normalized keys:
\begin{equation*}
    k_{p,h}^{\mathrm{wnp}}
    =
    P_h k_{p,h}^{\mathrm{wn}}
    =
    P_h W_h^{1/2} k_{p,h}/s_p.
\end{equation*}

\paragraph{Empirical analysis.}
Figure~\ref{fig:method-validation}(b--c) examines whether the covariance-aware
cost serves as an effective proxy for actual VQ distortion.
Despite depending only on group covariance, the proposed cost closely tracks
the reconstruction MSE, with a Pearson correlation of $r=0.994$ in the
representative layer and consistently strong correlations across layers.
Moreover, Figure~\ref{fig:method-validation}(d) shows that optimizing this
cost through covariance-aware grouping reduces both the grouping objective
and the resulting VQ reconstruction error relative to contiguous grouping.

\subsection{Vector Quantization in the Tailored Quantization Space}
\label{sec:method:encode}

We apply VQ to the keys $k^{\mathrm{wnp}}$ lying in the weighted, normalized, and grouped quantization space.
A separate $K$-centroid codebook $C_{h,G}$ is fitted for each layer, KV head, and channel group.

\paragraph{Fisher-weighted codebook fitting.}
While channel weighting captures \textit{channel-wise} importance, calibration tokens can also differ in their influence on the language-modeling loss.
To account for this \textit{token-wise} variation, we adopt Fisher-weighted k-means following \citet{zhang2024kv, kim2023squeezellm}.
Let $F_{p,h,i}=(\partial L/\partial k^{\mathrm{wnp}}_{p,h,i})^2$ denote the diagonal empirical Fisher, and let $\mathcal{G}$ denote the set of channel groups.
We assign each group $G\in\mathcal{G}$ a Fisher weight by summing over its channels:
\begin{equation*}
    \omega_{p,h,G}
    =
    \sum_{i\in G}F_{p,h,i}.
\end{equation*}
We then fit the codebook $C_{h,G}$ using the calibration dataset with the weighted k-means objective:
\begin{equation*}
    C_{h,G}
    =
    \operatorname*{argmin}_{C}
    \sum_p
    \omega_{p,h,G}
    \left\lVert
        k^{\mathrm{wnp}}_{p,h,G}
        -
        c_{z_{h,G}(k^{\mathrm{wnp}}_{p,h,G})}
    \right\rVert_2^2.
\end{equation*}

At inference, each group is encoded by its nearest codeword,
$z_{p,h,G}=\arg\min_j
\lVert k^{\mathrm{wnp}}_{p,h,G}-c_{h,G,j}\rVert_2^2$.
We store the resulting indices together with the shared scale $s_p$.

\paragraph{Runtime reconstruction.}
Let $W_{K,h}$ and $W_{Q,h}$ denote the key and associated query projection matrices, respectively.
We absorb the weighting and permutation into the key projection,
$\widetilde{W}_{K,h}=P_h W_h^{1/2} W_{K,h}$,
and the inverse weighting into the decoder codebooks by transforming each
codeword as
\[
\widetilde{c}_{h,G,j}
=
\left[
\left(P_h W_h P_h^\top\right)^{-1/2}
\right]_{G,G}
c_{h,G,j}.
\]
We denote the resulting collection of transformed codebooks by
$\widetilde{C}_h$.
Given the stored VQ indices $z_{p,h}$ and shared token scale $s_p$,
we reconstruct the key and compute the query--key dot product as
\begin{equation*}
\hat{k}^{p}_{p,h}
=
s_p \widetilde{C}_{h}[z_{p,h}],
\qquad
\ell_{t,p,h}
=
\left(
\widetilde{R}_{t,h}\widetilde{W}_{Q,h}x_t
\right)^\top
\widetilde{R}_{p,h}\hat{k}^{p}_{p,h},
\end{equation*}
where
$\widetilde{W}_{Q,h}=P_hW_{Q,h}$ and
$\widetilde{R}_{p,h}=P_hR_pP_h^\top$.
Thus, both keys and queries remain in the permuted coordinate system throughout attention.
Because $P_h$ permutes complete RoPE pairs, $\widetilde{R}_{p,h}$ retains the standard block-diagonal RoPE structure; implementing it only requires reordering the RoPE frequencies according to the permuted pair layout.

The modified projection weights and decoder codebooks are materialized offline.
During attention, codeword lookup, scale restoration, pairwise RoPE, and the query--key dot product are fused into the attention kernel.
Consequently, \method{} requires no runtime channel-shuffling operation, dense transformation, or additional kernel launch over standard VQ inference.

\paragraph{Value path.}
For values, we retain standard VQ with contiguous channel groups.
As shown in Figure~\ref{fig:key-value-channel-structure}, values exhibit substantially less pronounced channel-wise characteristics than keys, making channel weighting and grouping less beneficial.
Further analysis and value-side alternatives are provided in Appendix~\ref{app:value-design}.


\begin{table}[t!]
\centering
\caption{General performance under low-bit KV cache quantization. All results use greedy decoding. Bold denotes the best quantized result; higher is better.}
\label{tab:non_reasoning_bench}
\resizebox{\linewidth}{!}{
\begin{tabular}{llcrrrrrrr}
\toprule
Model & Method & Bits (K/V) & GSM8K & MATH500 & MBPP & HumanEval & BBH & MMLU & Avg. \\
\midrule
Llama-3.1-8B-Instruct & BF16 & 16.000/16.000 & 83.62 & 42.20 & 59.60 & 62.20 & 73.38 & 62.64 & 63.94 \\
\cmidrule(lr){2-10}
 & CQ & 1.250/1.250 & 68.01 & 19.60 & 52.00 & 54.27 & 40.08 & 54.21 & 48.03 \\
 & NovaKV & 1.375/1.250 & 76.95 & 29.40 & 55.00 & 53.66 & 48.46 & 56.43 & 53.32 \\
 & NSNQuant & 1.238/1.238 & 73.39 & 31.40 & 50.20 & 57.93 & 48.57 & \textbf{58.33} & 53.30 \\
 & \cellcolor{gray!15}\method{} & \cellcolor{gray!15}1.266/1.250 & \cellcolor{gray!15}\textbf{81.73} & \cellcolor{gray!15}\textbf{34.00} & \cellcolor{gray!15}\textbf{58.40} & \cellcolor{gray!15}\textbf{58.54} & \cellcolor{gray!15}\textbf{64.28} & \cellcolor{gray!15}\textbf{58.33} & \cellcolor{gray!15}\textbf{59.21} \\
\midrule
Qwen3-4B & BF16 & 16.000/16.000 & 86.28 & 72.60 & 64.00 & 81.71 & 78.03 & 74.72 & 76.22 \\
\cmidrule(lr){2-10}
 & CQ & 1.250/1.250 & 70.36 & 60.80 & 45.40 & 65.85 & 52.08 & 64.14 & 59.77 \\
 & NovaKV & 1.375/1.250 & 84.53 & 66.80 & \textbf{62.60} & 76.22 & \textbf{68.80} & 67.21 & 71.03 \\
 & NSNQuant & 1.238/1.238 & 65.58 & 52.40 & 46.20 & 67.68 & 49.57 & 63.29 & 57.45 \\
 & \cellcolor{gray!15}\method{} & \cellcolor{gray!15}1.266/1.250 & \cellcolor{gray!15}\textbf{85.44} & \cellcolor{gray!15}\textbf{69.80} & \cellcolor{gray!15}\textbf{62.60} & \cellcolor{gray!15}\textbf{77.44} & \cellcolor{gray!15}68.28 & \cellcolor{gray!15}\textbf{71.26} & \cellcolor{gray!15}\textbf{72.47} \\
\bottomrule
\end{tabular}
}
\end{table}

\begin{table}[t!]
\centering
\caption{Long-CoT reasoning performance under low-bit KV cache quantization. We report the mean and standard deviation over three seeds. Bold denotes the best quantized result; higher is better.}
\label{tab:reasoning_bench}
\resizebox{\linewidth}{!}{
\begin{tabular}{llcrrrrr}
\toprule
Model & Method & Bits (K/V) & AIME'24 & AIME'25 & LCB-v6 & SciBench & Avg. \\
\midrule
Qwen3-4B-Thinking-2507 & BF16 & 16.000/16.000 & 74.44$\pm$1.92 & 74.44$\pm$5.09 & 45.85$\pm$0.62 & 73.99$\pm$0.43 & 67.18$\pm$1.37 \\
\cmidrule(lr){2-8}
 & CQ & 1.250/1.250 & 26.67$\pm$3.33 & 14.44$\pm$5.09 & 18.99$\pm$0.44 & 54.00$\pm$0.94 & 28.52$\pm$1.54 \\
 & NovaKV & 1.375/1.250 & 7.78$\pm$1.92 & 6.67$\pm$3.33 & 9.16$\pm$0.38 & 27.94$\pm$2.08 & 12.89$\pm$1.10 \\
 & NSNQuant & 1.238/1.238 & 44.44$\pm$6.94 & 38.89$\pm$1.92 & 34.66$\pm$0.33 & 68.69$\pm$0.55 & 46.67$\pm$1.81 \\
 & \cellcolor{gray!15}\method{} & \cellcolor{gray!15}1.266/1.250 & \cellcolor{gray!15}\textbf{68.89$\pm$1.92} & \cellcolor{gray!15}\textbf{56.67$\pm$3.33} & \cellcolor{gray!15}\textbf{43.00$\pm$0.47} & \cellcolor{gray!15}\textbf{73.65$\pm$0.60} & \cellcolor{gray!15}\textbf{60.55$\pm$0.98} \\
\midrule
DeepSeek-R1-Distill-Llama-8B & BF16 & 16.000/16.000 & 53.33$\pm$3.33 & 31.11$\pm$1.92 & 39.68$\pm$0.56 & 38.49$\pm$0.30 & 40.65$\pm$0.98 \\
\cmidrule(lr){2-8}
 & CQ & 1.250/1.250 & 26.67$\pm$3.33 & 26.67$\pm$5.77 & 23.29$\pm$0.45 & 35.45$\pm$1.61 & 28.02$\pm$1.72 \\
 & NovaKV & 1.375/1.250 & 35.56$\pm$5.09 & 18.89$\pm$3.85 & 20.98$\pm$0.65 & 33.62$\pm$0.44 & 27.26$\pm$1.61 \\
 & NSNQuant & 1.238/1.238 & 44.44$\pm$5.09 & 24.44$\pm$5.09 & 32.67$\pm$1.15 & 34.83$\pm$1.64 & 34.10$\pm$1.87 \\
 & \cellcolor{gray!15}\method{} & \cellcolor{gray!15}1.266/1.250 & \cellcolor{gray!15}\textbf{48.89$\pm$6.94} & \cellcolor{gray!15}\textbf{31.11$\pm$1.92} & \cellcolor{gray!15}\textbf{32.95$\pm$0.61} & \cellcolor{gray!15}\textbf{39.11$\pm$1.50} & \cellcolor{gray!15}\textbf{38.02$\pm$1.85} \\
\midrule
Phi4-14B-Reasoning-Plus & BF16 & 16.000/16.000 & 71.11$\pm$1.92 & 66.67$\pm$3.33 & 46.60$\pm$0.71 & 48.22$\pm$1.03 & 58.15$\pm$1.01 \\
\cmidrule(lr){2-8}
 & CQ & 1.250/1.250 & 52.22$\pm$3.85 & 31.11$\pm$5.09 & 13.33$\pm$0.55 & 33.14$\pm$1.34 & 32.45$\pm$1.64 \\
 & NovaKV & 1.375/1.250 & 41.11$\pm$10.18 & 35.56$\pm$1.92 & 21.93$\pm$0.29 & 41.57$\pm$1.86 & 35.04$\pm$2.63 \\
 & NSNQuant & 1.238/1.238 & 63.33$\pm$6.67 & 55.56$\pm$5.09 & 37.12$\pm$1.04 & 44.99$\pm$0.65 & 50.25$\pm$2.12 \\
 & \cellcolor{gray!15}\method{} & \cellcolor{gray!15}1.263/1.250 & \cellcolor{gray!15}\textbf{71.11$\pm$5.09} & \cellcolor{gray!15}\textbf{60.00$\pm$0.00} & \cellcolor{gray!15}\textbf{41.20$\pm$0.52} & \cellcolor{gray!15}\textbf{50.58$\pm$1.90} & \cellcolor{gray!15}\textbf{55.72$\pm$1.36} \\
\bottomrule
\end{tabular}
}
\end{table}

\begin{table}[t]
\centering
\caption{Needle-in-a-haystack retrieval performance at increasing context lengths. We report the mean and standard deviation over three seeds. Bold denotes the best quantized result; higher is better.}
\label{tab:niah}
\resizebox{\linewidth}{!}{
\begin{tabular}{llcrrrrr}
\toprule
Model & Method & Bits (K/V) & 4k & 8k & 16k & 32k & 64k \\
\midrule
Llama-3.1-8B-Instruct & BF16 & 16.000/16.000 & 98.50$\pm$1.73 & 98.83$\pm$0.76 & 98.67$\pm$0.29 & 98.67$\pm$0.76 & 98.17$\pm$1.04 \\
\cmidrule(lr){2-8}
 & CQ & 1.250/1.250 & 19.33$\pm$1.61 & 15.50$\pm$3.61 & 12.67$\pm$1.26 & 12.83$\pm$3.18 & 6.33$\pm$1.26 \\
 & NovaKV & 1.375/1.250 & 84.83$\pm$0.29 & 81.67$\pm$3.18 & 64.00$\pm$2.18 & 21.33$\pm$2.25 & 1.83$\pm$0.29 \\
 & NSNQuant & 1.238/1.238 & 89.17$\pm$1.61 & 86.00$\pm$1.32 & 86.83$\pm$4.25 & 88.17$\pm$1.15 & 82.17$\pm$0.58 \\
 & \cellcolor{gray!15}\method{} & \cellcolor{gray!15}1.266/1.250 & \cellcolor{gray!15}\textbf{93.33$\pm$2.08} & \cellcolor{gray!15}\textbf{93.17$\pm$0.76} & \cellcolor{gray!15}\textbf{92.00$\pm$1.32} & \cellcolor{gray!15}\textbf{93.67$\pm$1.04} & \cellcolor{gray!15}\textbf{93.17$\pm$0.76} \\
\midrule
Qwen3-4B-Thinking-2507 & BF16 & 16.000/16.000 & 100.00$\pm$0.00 & 100.00$\pm$0.00 & 99.83$\pm$0.29 & 99.67$\pm$0.29 & 95.33$\pm$2.36 \\
\cmidrule(lr){2-8}
 & CQ & 1.250/1.250 & 71.50$\pm$1.73 & 63.33$\pm$4.01 & 48.17$\pm$0.29 & 21.83$\pm$1.04 & 10.50$\pm$2.78 \\
 & NovaKV & 1.375/1.250 & \textbf{99.83$\pm$0.29} & 27.67$\pm$2.25 & 0.00$\pm$0.00 & 0.00$\pm$0.00 & 0.00$\pm$0.00 \\
 & NSNQuant & 1.238/1.238 & 97.83$\pm$2.47 & 97.00$\pm$0.50 & 93.67$\pm$0.76 & 91.17$\pm$0.29 & 73.00$\pm$2.18 \\
 & \cellcolor{gray!15}\method{} & \cellcolor{gray!15}1.266/1.250 & \cellcolor{gray!15}\textbf{99.83$\pm$0.29} & \cellcolor{gray!15}\textbf{98.67$\pm$0.76} & \cellcolor{gray!15}\textbf{98.83$\pm$0.58} & \cellcolor{gray!15}\textbf{98.50$\pm$0.00} & \cellcolor{gray!15}\textbf{81.50$\pm$1.50} \\
\midrule
Phi4-14B-Reasoning-Plus & BF16 & 16.000/16.000 & 99.83$\pm$0.29 & 99.83$\pm$0.29 & 99.67$\pm$0.29 & 99.50$\pm$0.00 & -- \\
\cmidrule(lr){2-8}
 & CQ & 1.250/1.250 & 62.00$\pm$1.32 & 54.83$\pm$3.33 & 47.67$\pm$3.51 & 34.33$\pm$4.25 & -- \\
 & NovaKV & 1.375/1.250 & 97.33$\pm$1.26 & 91.83$\pm$1.61 & 62.50$\pm$3.28 & 0.00$\pm$0.00 & -- \\
 & NSNQuant & 1.238/1.238 & 98.67$\pm$0.76 & 98.17$\pm$1.44 & 98.00$\pm$0.50 & 92.67$\pm$1.26 & -- \\
 & \cellcolor{gray!15}\method{} & \cellcolor{gray!15}1.263/1.250 & \cellcolor{gray!15}\textbf{99.83$\pm$0.29} & \cellcolor{gray!15}\textbf{99.33$\pm$0.76} & \cellcolor{gray!15}\textbf{98.67$\pm$0.58} & \cellcolor{gray!15}\textbf{97.33$\pm$0.76} & \cellcolor{gray!15}-- \\
\bottomrule
\end{tabular}
}
\end{table}

\section{Experiments}
\label{sec:exp_results}
\subsection{Experimental Setup}
\label{sec:setup}

\paragraph{Models.}
We evaluate Llama-3.1-8B-Instruct \citep{grattafiori2024llama}, Qwen3-4B,
Qwen3-4B-Thinking-2507 \citep{yang2025qwen3}, Deepseek-R1-Distill-Llama-8B \citep{guo2025deepseek} and Phi-4-reasoning-plus
\citep{abdin2025phi}.  The first two models are used for general evaluation, and the
latter three are used for reasoning.  Long-context retrieval uses Llama-3.1-8B-Instruct, Qwen3-4B-Thinking-2507 and Phi-4-reasoning-plus.

\paragraph{Benchmarks.}
We evaluate general performance on GSM8K \citep{cobbe2021training}, MATH500
\citep{lightman2024let}, MBPP \citep{austin2021program}, HumanEval
\citep{chen2021evaluating}, BBH \citep{suzgun2023challenging} and MMLU \citep{hendrycks2020measuring} with greedy decoding.
Reasoning performance is evaluated on AIME~2024, AIME~2025, LiveCodeBench v6 \citep{jain2025livecodebench},
and SciBench \citep{wang2023scibench} with thinking enabled and three sampling seeds; we report mean
and standard deviation.  
Long-context
retrieval is evaluated with RULER \citep{hsieh2024ruler} needle-in-a-haystack at 4k, 8k, 16k, 32k,
and 64k using 200 examples and three seeds per setting.

\paragraph{Calibration.}
CQ \citep{zhang2024kv}, NovaKV \citep{fernandez2026spend}, and \method{} use the same 64
calibration windows of 2,048 tokens: 48 from \texttt{gpqa\_main} \citep{rein2023gpqa} and 16 from
\texttt{codeparrot/codeparrot-clean} \citep{codeparrot2022clean}.
The same windows are used for activation collection and, for CQ and \method{}, Fisher estimation.
NSNQuant \citep{son2026nsnquant} uses its calibration-free global codebook.

\paragraph{Decoding and residual policy.}
General tasks use greedy decoding and retain the most recent 64 tokens in full precision.
Reasoning tasks use temperature $0.6$, top-$p$ $0.95$, and generation limits up to 32,768 tokens;
the first 64 and most recent 256 tokens remain in full precision. For reasoning and retrieval tasks, we report the mean and standard deviation over three sampling seeds. Each benchmark uses the same
policy for all quantized methods.

\paragraph{Baselines and bit accounting.}
We compare \method{} with the pre-RoPE VQ baseline CQ and the post-RoPE VQ baselines NSNQuant and NovaKV.  We use the
published 1-bit-regime configurations of CQ and NSNQuant.  Because NovaKV does not report a
corresponding configuration, we retain its original quantizer design and match its K codebook to
the 1,024 centroids used by CQ and \method{}. The resulting K/V rates are reported in each result table. Appendix~\ref{app:bit-accounting}
provides the full accounting.

\paragraph{Implementation and hardware.}
We implement all methods in SGLang \citep{zheng2024sglang} with Triton attention kernels. All efficiency experiments are run on NVIDIA RTX 6000 Ada GPUs.

\subsection{Experiment Results}

Tables~\ref{tab:non_reasoning_bench}, \ref{tab:reasoning_bench}, and \ref{tab:niah} summarize the main results. On general tasks, \method{} achieves the strongest average performance across all models and leads on nearly every benchmark, substantially narrowing the gap to BF16. The improvement becomes more pronounced on long-CoT reasoning, where \method{} preserves considerably more of the full-precision performance than existing VQ baselines. A similar trend appears in long-context retrieval: as the context length
increases, \method{} exhibits substantially less degradation.
Interestingly, while NovaKV performs fairly well on  general tasks, it suffers from severe performance degradation on long-context benchmarks.
Appendix~\ref{app:position-generalization} further analyzes this trend at the attention-logit level beyond the calibration length.
\begin{figure}[t]
    \centering
    \includegraphics[width=0.9\linewidth]{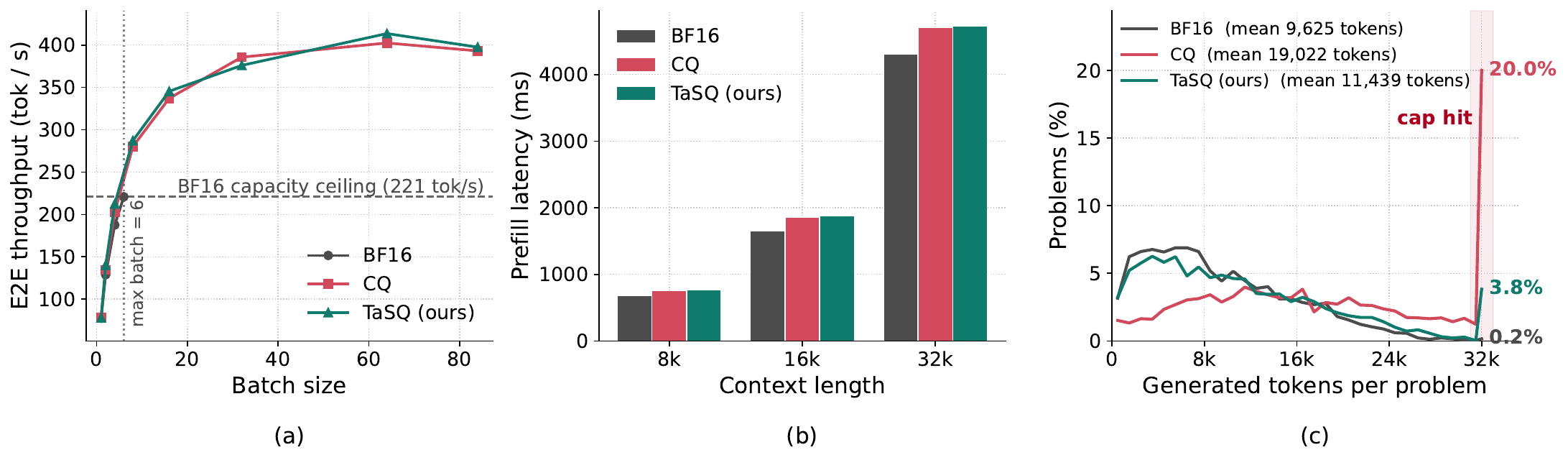}
     \caption{Serving efficiency of Qwen3-4B-Thinking-2507 model measured on one RTX 6000 Ada GPU.
  \textbf{(a)} End-to-end throughput vs.\ batch size (2048 in / 32768 out); dashed
  line = BF16 peak, dotted line = BF16's largest feasible batch.
  \textbf{(b)} Batch-1 prefill latency at 8k/16k/32k.
  \textbf{(c)} Generated tokens per problem on LiveCodeBench-v6, with the 32k cap
  shaded and the cap-hit rate per method.}

    \label{fig:discussion-efficiency-analysis}
\end{figure}

\section{Discussion}
\subsection{Efficiency Analysis}

\paragraph{Serving efficiency.} 
We compare the serving efficiency of \method{} with the BF16 baseline and CQ, a minimal KV cache VQ baseline without additional runtime transformations.
All three are evaluated within the same SGLang serving stack on a single RTX 6000 Ada GPU.
As shown in Figure~\ref{fig:discussion-efficiency-analysis}(a), \method{} matches the throughput of CQ, indicating negligible additional serving overhead.  Compared with BF16, it expands the
KV cache pool from \(235{,}257\) to \(2{,}935{,}184\) tokens and increases the maximum batch size
from 6 to 84.
This raises peak throughput from 220.8 to 412.6 tokens/s, a \(1.87\times\)
improvement.
Figure~\ref{fig:discussion-efficiency-analysis}(b) shows that VQ encode overhead increases time-to-first-token (TTFT) by 10\% to 14\% for 8k to 32k prompts, but this one-time prefill cost is amortized over long generations.

\paragraph{Reasoning stability.}
Low-bit KV cache compression can destabilize reasoning models, causing repetitive or incoherent
generation that continues until the length limit \citep{cheong2026kv}.
Figure~\ref{fig:discussion-efficiency-analysis}(c) shows that \method{} maintains output lengths and
cap-hit rates close to BF16.  Avoiding cap hits improves both accuracy and efficiency because
unterminated generations mostly fail the task while consuming the full decoding budget.  \method{}
therefore preserves reasoning stability while retaining the serving benefits of low-bit
KV cache compression.
\subsection{Ablation Study}

\begin{table}[t]
\centering
\caption{Component ablation of \method{} on base Llama-3.1-8B. We report WikiText-2 perplexity at sequence length
2,048.
Bold denotes the best quantized result.}
\label{tab:component-ablation}
\resizebox{0.4\columnwidth}{!}{%
  \begin{tabular}{lc}
\toprule
Method & PPL $\downarrow$ \\
\midrule
FP16 & 6.2374 \\
\midrule
Full \method{} & \textbf{8.4559} \\
\quad w/o query-guided weighting & 8.5172 \\
\quad w/o cross-head normalization & 8.5103 \\
\quad w/o covariance-aware grouping & 10.1731 \\
\bottomrule
\end{tabular}
}
\end{table}

\begin{figure}[t]
    \centering
    \includegraphics[width=\linewidth]{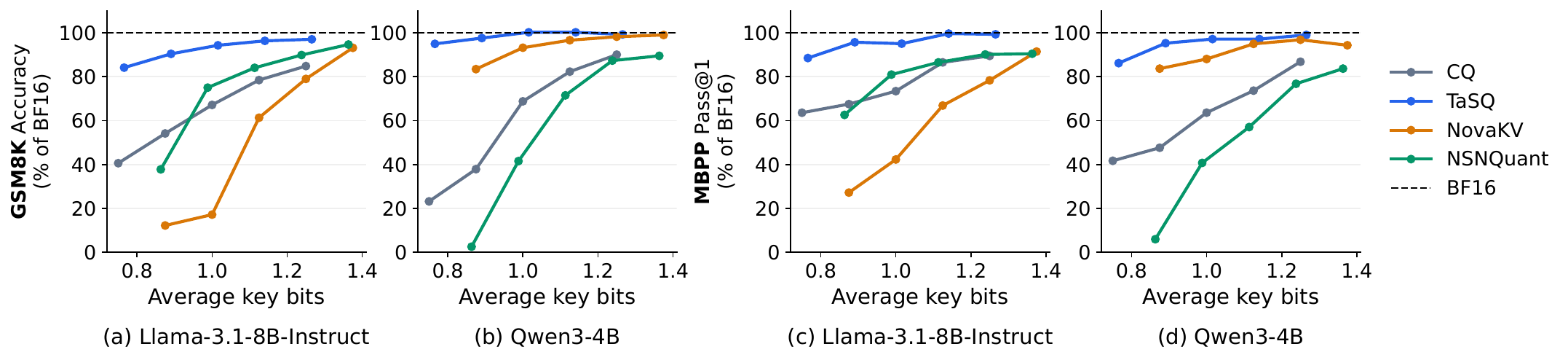}
    \vspace{-1.5em}
    \caption{GSM8K and MBPP performance versus average key bits for Llama-3.1-8B-Instruct (a, c) and Qwen3-4B (b, d). Scores are normalized to each model's BF16 baseline (dashed, 100\%); values remain in BF16.}
    \label{fig:gsm8k-average-key-bits}
\end{figure}

\paragraph{Component ablation.} To assess the contribution of individual method components, we measure perplexity (PPL) using Llama-3.1-8B on WikiText-2 with a context length of 2,048, while removing each component individually.
We use the same calibration settings as in the main experiments and do not preserve any portion of the KV cache in full precision.
The results in Table~\ref{tab:component-ablation} show that while all components improve quantization quality, removing grouping results in the largest increase in perplexity, suggesting that grouping makes the largest contribution among the three components.

\paragraph{Evaluation under different bitwidths.}
We further compare the rate--accuracy trade-off on GSM8K and MBPP using Llama-3.1-8B-Instruct and Qwen3-4B, evaluating each method over five codebook sizes spanning comparable key rates.
Since NovaKV uses scalar quantization for values, it cannot sweep the value rate into the sub-1-bit regime; we therefore keep values in BF16 for all methods and vary only the key rate.
Exact configurations and bit accounting are provided in Appendix~\ref{app:bit-accounting}.
As shown in Figure~\ref{fig:gsm8k-average-key-bits}, TaSQ consistently achieves the strongest rate--accuracy trade-off across both models, with the advantage becoming more pronounced at lower key rates.
\section{Conclusion}

We introduced \method{}, which tailors the quantization target space for accurate and efficient KV cache VQ in the 1-bit regime.
Our design incorporates query-guided weighting, cross-head normalization, and covariance-aware grouping while retaining the serving efficiency of conventional KV cache VQ methods.
\method{} consistently outperforms existing VQ baselines across
general, reasoning, and long-context retrieval tasks while preserving generation stability.
\method{} supports up to $14\times$ larger batches and achieves up to $1.87\times$ higher throughput than the BF16 baseline in our SGLang implementation.
Together, these results demonstrate the effectiveness of tailoring the VQ target space for ultra-low-bit KV cache compression.

\bibliography{iclr2027_conference}

@book{gersho2012vector,
  title={Vector quantization and signal compression},
  author={Gersho, Allen and Gray, Robert M},
  year={2012},
  publisher={Springer Science \& Business Media}
}

@article{zhang2024kv,
  title={Kv cache is 1 bit per channel: Efficient large language model inference with coupled quantization},
  author={Zhang, Tianyi and Yi, Jonah and Xu, Zhaozhuo and Shrivastava, Anshumali},
  journal={Advances in Neural Information Processing Systems},
  volume={37},
  pages={3304--3331},
  year={2024}
}

@article{son2026nsnquant,
  title={NSNQuant: A double normalization approach for calibration-free low-bit vector quantization of KV cache},
  author={Son, Donghyun and Choi, Euntae and Yoo, Sungjoo},
  journal={Advances in Neural Information Processing Systems},
  volume={38},
  pages={43124--43159},
  year={2026}
}

@article{fernandez2026spend,
  title={Spend Bits Where Queries Look: KV Cache Vector Quantization with Attention-Preserving Transforms},
  author={Fern{\'a}ndez-Mendui{\~n}a, Samuel and Ziashahabi, Amir and Pavez, Eduardo and Ortega, Antonio and Avestimehr, Salman},
  journal={arXiv preprint arXiv:2608.04074},
  year={2026}
}

@article{liu2024kivi,
  title={Kivi: A tuning-free asymmetric 2bit quantization for kv cache},
  author={Liu, Zirui and Yuan, Jiayi and Jin, Hongye and Zhong, Shaochen and Xu, Zhaozhuo and Braverman, Vladimir and Chen, Beidi and Hu, Xia},
  journal={arXiv preprint arXiv:2402.02750},
  year={2024}
}

@article{hooper2024kvquant,
  title={Kvquant: Towards 10 million context length llm inference with kv cache quantization},
  author={Hooper, Coleman and Kim, Sehoon and Mohammadzadeh, Hiva and Mahoney, Michael W and Shao, Yakun S and Keutzer, Kurt and Gholami, Amir},
  journal={Advances in Neural Information Processing Systems},
  volume={37},
  pages={1270--1303},
  year={2024}
}

@article{grattafiori2024llama,
  title={The llama 3 herd of models},
  author={Grattafiori, Aaron and Dubey, Abhimanyu and Jauhri, Abhinav and Pandey, Abhinav and Kadian, Abhishek and Al-Dahle, Ahmad and Letman, Aiesha and Mathur, Akhil and Schelten, Alan and Vaughan, Alex and others},
  journal={arXiv preprint arXiv:2407.21783},
  year={2024}
}

@article{yang2025qwen3,
  title={Qwen3 technical report},
  author={Yang, An and Li, Anfeng and Yang, Baosong and Zhang, Beichen and Hui, Binyuan and Zheng, Bo and Yu, Bowen and Gao, Chang and Huang, Chengen and Lv, Chenxu and others},
  journal={arXiv preprint arXiv:2505.09388},
  year={2025}
}

@article{abdin2025phi,
  title={Phi-4-reasoning technical report},
  author={Abdin, Marah and Agarwal, Sahaj and Awadallah, Ahmed and Balachandran, Vidhisha and Behl, Harkirat and Chen, Lingjiao and de Rosa, Gustavo and Gunasekar, Suriya and Javaheripi, Mojan and Joshi, Neel and others},
  journal={arXiv preprint arXiv:2504.21318},
  year={2025}
}

@article{cobbe2021training,
  title={Training verifiers to solve math word problems},
  author={Cobbe, Karl and Kosaraju, Vineet and Bavarian, Mohammad and Chen, Mark and Jun, Heewoo and Kaiser, Lukasz and Plappert, Matthias and Tworek, Jerry and Hilton, Jacob and Nakano, Reiichiro and others},
  journal={arXiv preprint arXiv:2110.14168},
  year={2021}
}

@inproceedings{lightman2024let,
  title={Let's verify step by step},
  author={Lightman, Hunter and Kosaraju, Vineet and Burda, Yuri and Edwards, Harrison and Baker, Bowen and Lee, Teddy and Leike, Jan and Schulman, John and Sutskever, Ilya and Cobbe, Karl},
  booktitle={International Conference on Learning Representations},
  volume={2024},
  pages={39578--39601},
  year={2024}
}

@article{austin2021program,
  title={Program synthesis with large language models},
  author={Austin, Jacob and Odena, Augustus and Nye, Maxwell and Bosma, Maarten and Michalewski, Henryk and Dohan, David and Jiang, Ellen and Cai, Carrie and Terry, Michael and Le, Quoc and others},
  journal={arXiv preprint arXiv:2108.07732},
  year={2021}
}

@article{chen2021evaluating,
  title={Evaluating large language models trained on code},
  author={Chen, Mark and Tworek, Jerry and Jun, Heewoo and Yuan, Qiming and Pinto, Henrique Ponde De Oliveira and Kaplan, Jared and Edwards, Harri and Burda, Yuri and Joseph, Nicholas and Brockman, Greg and others},
  journal={arXiv preprint arXiv:2107.03374},
  year={2021}
}

@inproceedings{suzgun2023challenging,
  title={Challenging big-bench tasks and whether chain-of-thought can solve them},
  author={Suzgun, Mirac and Scales, Nathan and Sch{\"a}rli, Nathanael and Gehrmann, Sebastian and Tay, Yi and Chung, Hyung Won and Chowdhery, Aakanksha and Le, Quoc and Chi, Ed H and Zhou, Denny and others},
  booktitle={Findings of the Association for Computational Linguistics: ACL 2023},
  pages={13003--13051},
  year={2023}
}

@article{jain2025livecodebench,
  title={LiveCodeBench: Holistic and Contamination Free Evaluation of Large Language Models for Code},
  author={Naman Jain and King Han and Alex Gu and Wen-Ding Li and Fanjia Yan and Tianjun Zhang and Sida Wang and Armando Solar-Lezama and Koushik Sen and Ion Stoica},
  journal={ArXiv},
  year={2024},
  volume={abs/2403.07974},
  url={https://api.semanticscholar.org/CorpusID:268379413}
}

@article{rein2023gpqa,
  title={Gpqa: A graduate-level google-proof q\&a benchmark},
  author={Rein, David and Hou, Betty Li and Stickland, Asa Cooper and Petty, Jackson and Pang, Richard Yuanzhe and Dirani, Julien and Michael, Julian and Bowman, Samuel R},
  journal={arXiv preprint arXiv:2311.12022},
  year={2023}
}

@article{hsieh2024ruler,
  title={RULER: What's the real context size of your long-context language models?},
  author={Hsieh, Cheng-Ping and Sun, Simeng and Kriman, Samuel and Acharya, Shantanu and Rekesh, Dima and Jia, Fei and Zhang, Yang and Ginsburg, Boris},
  journal={arXiv preprint arXiv:2404.06654},
  year={2024}
}

@misc{codeparrot2022clean,
  title={CodeParrot Clean},
  author={{CodeParrot}},
  year={2022},
  howpublished={Hugging Face dataset},
  url={https://huggingface.co/datasets/codeparrot/codeparrot-clean}
}

@article{zheng2024sglang,
  title={Sglang: Efficient execution of structured language model programs},
  author={Zheng, Lianmin and Yin, Liangsheng and Xie, Zhiqiang and Sun, Chuyue and Huang, Jeff and Yu, Cody H and Cao, Shiyi and Kozyrakis, Christos and Stoica, Ion and Gonzalez, Joseph E and others},
  journal={Advances in neural information processing systems},
  volume={37},
  pages={62557--62583},
  year={2024}
}

@article{cheong2026kv,
  title={KV-Rescue: Recovering Reasoning Language Model KV Eviction Loss via Stepwise Interleaving},
  author={Cheong, Minsoo and Lim, Woosang and Yun, Vincent-Daniel and Yoo, Sungjoo},
  journal={arXiv preprint arXiv:2608.15797},
  year={2026}
}

@inproceedings{xu2025think,
  title={Think: Thinner key cache by query-driven pruning},
  author={Xu, Yuhui and Jie, Zhanming and Dong, Hanze and Wang, Lei and Lu, Xudong and Zhou, Aojun and Saha, Amrita and Xiong, Caiming and Sahoo, Doyen},
  booktitle={International Conference on Learning Representations},
  volume={2025},
  pages={56691--56709},
  year={2025}
}

@article{vaswani2017attention,
  title={Attention is all you need},
  author={Vaswani, Ashish and Shazeer, Noam and Parmar, Niki and Uszkoreit, Jakob and Jones, Llion and Gomez, Aidan N and Kaiser, {\L}ukasz and Polosukhin, Illia},
  journal={Advances in neural information processing systems},
  volume={30},
  year={2017}
}

@article{kim2023squeezellm,
  title={Squeezellm: Dense-and-sparse quantization},
  author={Kim, Sehoon and Hooper, Coleman and Gholami, Amir and Dong, Zhen and Li, Xiuyu and Shen, Sheng and Mahoney, Michael W and Keutzer, Kurt},
  journal={arXiv preprint arXiv:2306.07629},
  year={2023}
}

@article{frantar2022gptq,
  title={Gptq: Accurate post-training quantization for generative pre-trained transformers},
  author={Frantar, Elias and Ashkboos, Saleh and Hoefler, Torsten and Alistarh, Dan},
  journal={arXiv preprint arXiv:2210.17323},
  year={2022}
}

@article{lin2024awq,
  title={Awq: Activation-aware weight quantization for on-device llm compression and acceleration},
  author={Lin, Ji and Tang, Jiaming and Tang, Haotian and Yang, Shang and Chen, Wei-Ming and Wang, Wei-Chen and Xiao, Guangxuan and Dang, Xingyu and Gan, Chuang and Han, Song},
  journal={Proceedings of machine learning and systems},
  volume={6},
  pages={87--100},
  year={2024}
}

@inproceedings{xiao2023smoothquant,
  title={Smoothquant: Accurate and efficient post-training quantization for large language models},
  author={Xiao, Guangxuan and Lin, Ji and Seznec, Mickael and Wu, Hao and Demouth, Julien and Han, Song},
  booktitle={International conference on machine learning},
  pages={38087--38099},
  year={2023},
  organization={PMLR}
}

@article{van2024gptvq,
  title={Gptvq: The blessing of dimensionality for llm quantization},
  author={Van Baalen, Mart and Kuzmin, Andrey and Koryakovskiy, Ivan and Nagel, Markus and Couperus, Peter and Bastoul, Cedric and Mahurin, Eric and Blankevoort, Tijmen and Whatmough, Paul},
  journal={arXiv preprint arXiv:2402.15319},
  year={2024}
}

@article{egiazarian2024extreme,
  title={Extreme compression of large language models via additive quantization},
  author={Egiazarian, Vage and Panferov, Andrei and Kuznedelev, Denis and Frantar, Elias and Babenko, Artem and Alistarh, Dan},
  journal={arXiv preprint arXiv:2401.06118},
  year={2024}
}

@article{tseng2024quip,
  title={Quip\#: Even better llm quantization with hadamard incoherence and lattice codebooks},
  author={Tseng, Albert and Chee, Jerry and Sun, Qingyao and Kuleshov, Volodymyr and De Sa, Christopher},
  journal={Proceedings of machine learning research},
  volume={235},
  pages={48630},
  year={2024}
}

@article{duanmu2024skvq,
  title={Skvq: Sliding-window key and value cache quantization for large language models},
  author={Duanmu, Haojie and Yuan, Zhihang and Li, Xiuhong and Duan, Jiangfei and Zhang, Xingcheng and Lin, Dahua},
  journal={arXiv preprint arXiv:2405.06219},
  year={2024}
}

@article{zhang2023h2o,
  title={H2o: Heavy-hitter oracle for efficient generative inference of large language models},
  author={Zhang, Zhenyu and Sheng, Ying and Zhou, Tianyi and Chen, Tianlong and Zheng, Lianmin and Cai, Ruisi and Song, Zhao and Tian, Yuandong and R{\'e}, Christopher and Barrett, Clark and others},
  journal={Advances in neural information processing systems},
  volume={36},
  pages={34661--34710},
  year={2023}
}

@article{li2024snapkv,
  title={Snapkv: Llm knows what you are looking for before generation},
  author={Li, Yuhong and Huang, Yingbing and Yang, Bowen and Venkitesh, Bharat and Locatelli, Acyr and Ye, Hanchen and Cai, Tianle and Lewis, Patrick and Chen, Deming},
  journal={Advances in Neural Information Processing Systems},
  volume={37},
  pages={22947--22970},
  year={2024}
}

@article{wang2024model,
  title={Model tells you where to merge: Adaptive kv cache merging for llms on long-context tasks},
  author={Wang, Zheng and Jin, Boxiao and Yu, Zhongzhi and Zhang, Minjia},
  journal={arXiv preprint arXiv:2407.08454},
  year={2024}
}

@article{wang2023scibench,
  title={Scibench: Evaluating college-level scientific problem-solving abilities of large language models},
  author={Wang, Xiaoxuan and Hu, Ziniu and Lu, Pan and Zhu, Yanqiao and Zhang, Jieyu and Subramaniam, Satyen and Loomba, Arjun R and Zhang, Shichang and Sun, Yizhou and Wang, Wei},
  journal={arXiv preprint arXiv:2307.10635},
  year={2023}
}

@article{guo2025deepseek,
  title={Deepseek-r1: Incentivizing reasoning capability in llms via reinforcement learning},
  author={Guo, Daya and Yang, Dejian and Zhang, Haowei and Song, Junxiao and Wang, Peiyi and Zhu, Qihao and Xu, Runxin and Zhang, Ruoyu and Ma, Shirong and Bi, Xiao and others},
  journal={arXiv preprint arXiv:2501.12948},
  year={2025}
}

@article{hendrycks2020measuring,
  title={Measuring massive multitask language understanding},
  author={Hendrycks, Dan and Burns, Collin and Basart, Steven and Zou, Andy and Mazeika, Mantas and Song, Dawn and Steinhardt, Jacob},
  journal={arXiv preprint arXiv:2009.03300},
  year={2020}
}
\bibliographystyle{iclr2027_conference}

\clearpage

\appendix
\section{Related Work}
\label{sec:related-work}

LLM quantization initially focused on model weights. Scalar quantization (SQ), which quantizes
individual weights independently, has been widely studied~\citep{frantar2022gptq,lin2024awq,xiao2023smoothquant}.
At ultra-low bit rates, vector quantization (VQ) has gained attention for jointly encoding weight
blocks and offering more favorable rate--distortion tradeoffs~\citep{van2024gptvq,egiazarian2024extreme,tseng2024quip}.

KV cache quantization has emerged more recently as long contexts and larger serving batches shift
the memory bottleneck from static weights to dynamically growing activations. Existing approaches
span both SQ~\citep{liu2024kivi,hooper2024kvquant,duanmu2024skvq} and
VQ~\citep{zhang2024kv,son2026nsnquant}.
Concurrent work NovaKV adopts a hybrid design, using dense transform-based VQ for post-RoPE keys and rotation-based SQ for values \citep{fernandez2026spend}. Our work belongs to the KV cache VQ line and studies the design of its native-basis pre-RoPE target space.

\section{Baseline Configurations and Bit Accounting}
\label{app:bit-accounting}



We report the effective KV cache rate in bits per element, accounting for both
codebook indices and per-token metadata.
All VQ methods use eight-channel groups, resulting in an index cost of
$\log_2 K / 8$ bits per element for a codebook with $K$ centroids.
All evaluated models have head dimension $d=128$ and $H=8$ except Phi4-14B-Reasoning-Plus model, which has $H=10$ KV
heads. We exclude static codebook storage and the full-precision prefix and
recent windows, which are shared across methods.

For CQ and NSNQuant, we use their published configurations for the 1-bit
regime (i.e., CQ-8c10b and NSNQuant-1b). Since NovaKV does not provide a configuration for this regime, we retain its original key VQ and rotation-based value SQ designs while setting
the key codebook size to 1,024 centroids, matching CQ and TaSQ. Table~\ref{tab:bit-accounting-app} 
summarizes the resulting effective cache rates and their components.

\paragraph{Bitwidth sweep configurations.}
For the rate--accuracy sweep on GSM8K and MBPP in Figure~\ref{fig:gsm8k-average-key-bits}, we vary the key codebook size over
$K\in\{64,128,256,512,1024\}$ for CQ, NovaKV, and TaSQ.
For NSNQuant, which incurs additional normalization metadata, we instead use
$K\in\{32,64,128,256,512\}$ to cover a comparable range of effective key rates.
Values are kept in BF16 for all methods because NovaKV uses scalar quantization for values and therefore cannot cover the sub-1-bit value-rate regime.
Reported key rates include codebook indices and method-specific metadata, while excluding static codebook storage, full-precision cache windows, and implementation padding.

\begin{table*}[h]
\centering
\caption{Effective cache rate in bits per element.
Per-token metadata costs are included in the reported rates.}
\label{tab:bit-accounting-app}
\footnotesize
\setlength{\tabcolsep}{4pt}
\renewcommand{\arraystretch}{1.1}

\begin{tabularx}{\textwidth}{@{}l c X c@{}}
\toprule
Method & Channel & Representation & Rate \\
\midrule
CQ
& K & 1,024-centroid VQ
& \(10/8 = 1.250\) \\
& V & 1,024-centroid VQ
& \(10/8 = 1.250\) \\
\midrule

NSNQuant
& K & 256-centroid VQ + normalization metadata
& \(8/8 + 0.238 = 1.238\) \\
& V & 256-centroid VQ + normalization metadata
& \(8/8 + 0.238 = 1.238\) \\
\midrule

NovaKV
& K & 1,024-centroid VQ + one 16-bit head-wise scale
& \(10/8 + 16/128 = 1.375\) \\
& V & 1-bit SQ + two 16-bit head-wise parameters
& \(1 + 32/128 = 1.250\) \\
\midrule

\method{}
& K & 1,024-centroid VQ + one 16-bit cross-head scale
& \(10/8 + 16/(8\cdot128) = 1.266\) \\
& V & 1,024-centroid VQ
& \(10/8 = 1.250\) \\
\bottomrule
\end{tabularx}
\end{table*}

\section{Value Quantization Analysis}

\begin{figure}[!h]
    \centering
    \includegraphics[width=\linewidth]{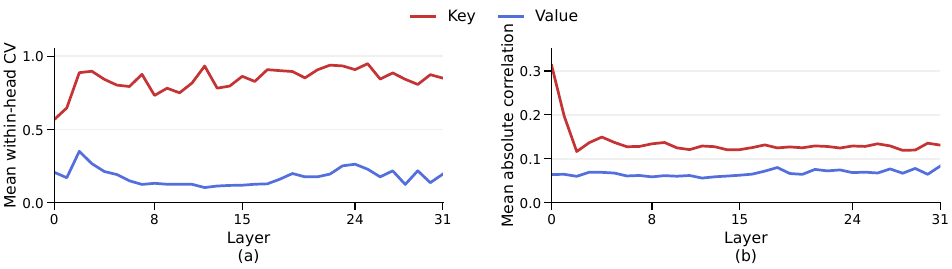}
    \setlength{\abovecaptionskip}{3pt}
    \caption{Key--value channel characteristics across layers of Llama-3.1-8B on the same
    2,048-token GPQA segment. (a) Mean coefficient of variation (CV) of channel
    sensitivity; lower values indicate flatter sensitivity. (b) Mean absolute
    inter-channel Pearson correlation with the diagonal excluded. Both metrics are
    averaged over the 8 KV heads.}
    \label{fig:key-value-channel-structure}
\end{figure}

\paragraph{Key--value channel characteristics.}
\label{app:key-value-structure}
We compare keys and values in terms of channel-wise sensitivity variation and
inter-channel correlation using a 2,048-token GPQA segment with Llama-3.1-8B.
We collect pre-RoPE keys, values, and their loss gradients across all 32 layers.

For $x\in\{k,v\}$, we measure the sensitivity of channel $i$ in head $h$ and
layer $l$ by its mean absolute loss gradient, and its variation by the coefficient of variation (CV):
\begin{equation*}
a^{x}_{l,h,i}
=
\frac{1}{T}\sum_{p=1}^{T}
\left|\frac{\partial L}{\partial x_{l,p,h,i}}\right|,
\qquad
\mathrm{CV}^{x}_{l}
=
\frac{1}{H}\sum_{h=1}^{H}
\frac{\operatorname{Std}_{i}(a^{x}_{l,h,i})}
{\operatorname{Mean}_{i}(a^{x}_{l,h,i})}.
\end{equation*}
Here, CV is computed within each head, with a lower CV indicating more uniform
sensitivity across channels. As shown in
Figure~\ref{fig:key-value-channel-structure}(a), values have substantially lower
CV than keys across all layers ($0.176$ vs.\ $0.841$ on average).
This more uniform sensitivity makes the original value space better aligned
with the uniform error metric of Euclidean VQ than the key space.

We also measure the mean absolute Pearson correlation between distinct channel
pairs within each head. Figure~\ref{fig:key-value-channel-structure}(b) shows
weaker inter-channel correlations for values across all layers
($0.067$ vs.\ $0.136$ on average).
Together, these results show that values exhibit both more uniform channel
sensitivity and weaker inter-channel dependencies than keys, motivating our
focus on tailoring the quantization space for keys.
We leave tailoring the quantization space for values to future work.

\paragraph{Explored value-side alternatives.}
\label{app:value-design}
Our final method leaves the value path unchanged and applies plain contiguous VQ.  As an exploratory
ablation, we tested whether the key-side weighting, normalization, and grouping choices also
benefit value quantization.
Following the derivation of query-guided key
weighting in Section~\ref{sec:query-guided-weighting}, we first derive an
analogous channel-wise sensitivity for values.
Let $\delta v_{p,h}=v_{p,h}-\hat{v}_{p,h}$ denote the value reconstruction
error at position $p$ and KV head $h$. The contribution of this error to
the attention output at query position $t$ is
\begin{equation*}
    \Delta o_{t,p,h}
    = a_{t,p,h} W_{O,h}\delta v_{p,h},
\end{equation*}
where $a_{t,p,h}$ is the attention weight and $W_{O,h}$ denotes the block
of the output projection corresponding to head $h$. Its squared norm is
\begin{equation*}
    \|\Delta o_{t,p,h}\|_2^2
    =
    \delta v_{p,h}^{\top}
    \left(a_{t,p,h}^2 W_{O,h}^{\top}W_{O,h}\right)
    \delta v_{p,h}.
\end{equation*}
Since $a_{t,p,h}^2$ scales all value channels equally, the relative
importance of value reconstruction errors is determined by
$W_{O,h}^\top W_{O,h}$. Following the diagonal approximation used for keys,
we therefore use $\operatorname{diag}(W_{O,h}^\top W_{O,h})$ to construct
channel weights for values.

Based on the derived weights, we evaluate the combined application of value-side channel weighting, cross-head normalization,
and covariance-aware grouping on Llama-3.1-8B while keeping the key path fixed.
We otherwise follow the main calibration and VQ settings, except that value grouping starts
from individual channels since values are not subject to RoPE-pair constraints.
Table~\ref{tab:value-side-ppl} shows that these transforms reduce WikiText-2 perplexity from
8.45594 to 8.42058 (0.42\%).
Given the modest gain and the additional bits that scales require, we retain standard VQ with contiguous channel
groups for values.

\begin{table}[t]
\centering
\caption{Value-side ablation on Llama-3.1-8B: WikiText-2 perplexity
at sequence length 2,048 (lower is better). The \method{} key path is
identical in both rows.}
\label{tab:value-side-ppl}
\small
\begin{tabular}{@{}lc@{}}
\toprule
Value quantization & PPL $\downarrow$ \\
\midrule
TaSQ & 8.45594 \\
TaSQ + value-side transformations & \textbf{8.42058} \\
\bottomrule
\end{tabular}
\end{table}

\begin{algorithm}[t]
\caption{RoPE-pair-constrained hierarchical partitioning}
\label{alg:pair-matching-app}
\begin{algorithmic}[1]
\Require covariance $\Sigma_h^{\mathrm{wn}}$;
group size $g$
\State $U\gets\{\{2j-1,2j\}:j=1,\ldots,d/2\}$
\For{$r=1$ \textbf{to} $\log_2(g/2)$}
    \State assign each edge $(u,v)$ between distinct units in $U$ the cost $c_h(u\cup v)$
    \State $M\gets$ minimum-weight perfect matching on $U$
    \State $U\gets\{u\cup v:(u,v)\in M\}$
\EndFor
\State \Return $U$ and the corresponding grouping layout
\end{algorithmic}
\end{algorithm}

\section{Method Details and Analysis}
\label{app:method-justification}
\paragraph{Hierarchical matching for covariance-aware grouping.}
\label{app:partition-algorithm}
Section~\ref{sec:channel-grouping} formulates covariance-aware grouping as
a partitioning problem that minimizes the total group cost subject to
equal-size groups and RoPE-pair preservation.
Let $n=d/2$ denote the number of atomic RoPE pairs and $m=g/2$ the number
of pairs per group. The number of possible partitions satisfying these
constraints is
\begin{equation*}
\frac{n!}{(m!)^{n/m}(n/m)!},
\end{equation*}
making exhaustive search infeasible.
We therefore approximately solve this problem using a hierarchical procedure
based on minimum-weight perfect matching, as detailed in
Algorithm~\ref{alg:pair-matching-app}.

Each round finds the exact minimum-cost pairing of its current groups.
Since earlier merges constrain the choices available in subsequent rounds,
the hierarchical procedure does not guarantee a globally optimal final
partition.

\paragraph{Cross-head shared scale.}
\label{app:shared-scale}
Section~\ref{sec:method:normalization} derives the metadata cost of cross-head normalization.
With a $b_s$-bit scale and head dimension $d$, head-wise normalization requires $b_s/d$ bits per
channel, whereas sharing the scale across $H$ heads reduces this cost to $b_s/(Hd)$. Under a fixed
bit budget, these savings can be reassigned to codebook indices, providing more centroids to
approximate each channel group and thereby reducing VQ distortion.

\begin{table}[h]
  \centering
  \small
  \begin{tabular}{lccccccc}
    \toprule
    & & \multicolumn{3}{c}{Llama-3.1-8B} & \multicolumn{3}{c}{Qwen3-4B} \\
    \cmidrule(lr){3-5} \cmidrule(lr){6-8}
    scale / codebook & bits/elem & mean & min & max & mean & min & max \\
    \midrule
    per-head, $K=1024$ & 1.3750 & 0.004483 & 0.002882 & 0.006953 & 0.019794 & 0.006732 & 0.030261 \\
    pooled, $K=1024$   & 1.2656 & 0.004564 & 0.002965 & 0.007037 & 0.019688 & 0.006754 & 0.030014 \\
    pooled, $K=2048$   & 1.3906 & 0.003460 & 0.002245 & 0.005313 & 0.014763 & 0.005045 & 0.022639 \\
    \bottomrule
  \end{tabular}
  \caption{Relative SSE with separately fitted codebooks, across all layers. Bits per channel
  include fp16 scale metadata with $g=8$, $d=128$, and $H=8$.}
  \label{tab:scale-rate-distortion-app}
\end{table}

At equal $K$, pooling is distortion-neutral: relative SSE stays within $\pm4\%$ of per-head on
every layer of both models, while saving $0.109$ bits per channel.  That is just short of the
$1/8$ bit a doubling of $K$ costs, so pooled $K=2048$ spends $0.016$ bits per channel more than
per-head $K=1024$; in exchange, relative SSE drops by at least $20\%$ in all $68$ layers measured.
With an 8-bit token scale and a shared block-level fp32 scale, pooling saves $0.0547$ bits per
channel for $H=8$ and $d=128$.

\begin{figure}[ht]
    \centering
    \includegraphics[width=\linewidth]{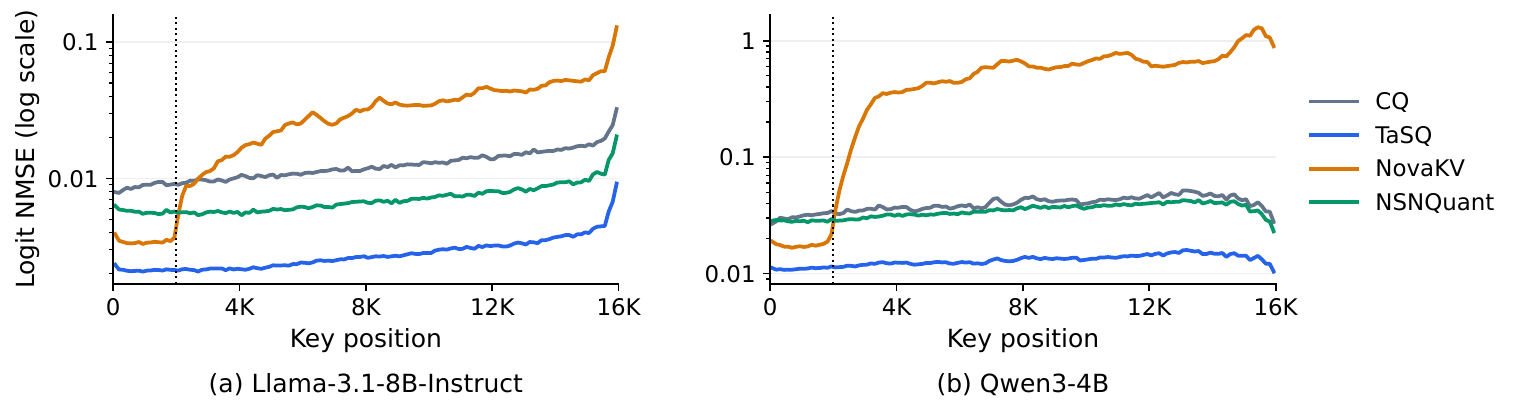}
    \caption{Attention-logit NMSE (log scale) versus key position for Llama-3.1-8B-Instruct (a) and Qwen3-4B (b).
    Dotted lines mark the 2048 calibration length.}
    \label{fig:key-error-position}
\end{figure}

\paragraph{Generalization beyond the calibration length.}
\label{app:position-generalization}
Because RoPE induces position-dependent changes in the key distribution, methods fitted to post-RoPE activations may be sensitive to positions beyond those observed during calibration.
To examine this effect, we evaluate all four methods on eight 16K WikiText-2 sequences per model, using the same calibration setting as in the main experiments.
Figure~\ref{fig:key-error-position} reports attention-logit NMSE over bins of 128 key positions, computed from causally valid query--key pairs across layers and heads.

For the pre-RoPE VQ methods, CQ and TaSQ, the error increases only gradually beyond the calibration range, with TaSQ maintaining the lowest error across positions.
NSNQuant, despite operating on post-RoPE keys, also remains relatively stable, due to its calibration-free codebook design.
In contrast, NovaKV exhibits a sharp increase in error beyond the calibration range.
Its transforms and codebooks are fitted directly to post-RoPE activations from the calibration data, making them more sensitive to position-dependent distribution shifts outside the observed range.
This trend is consistent with the long-context retrieval results in Table~\ref{tab:niah}, where NovaKV degrades sharply as context length increases, while TaSQ maintains substantially stronger performance.

\paragraph{Justification for the diagonal approximation in query-guided channel weighting.}
\label{app:diag-w}
Section~\ref{sec:query-guided-weighting} uses
$W_h=\operatorname{diag}(\tilde{A}_h)$ and transforms keys with $W_h^{1/2}$ rather than the dense
$\tilde{A}_h^{1/2}$. We justify this choice by comparing decode cost and evaluation error.

Let $u_{p,h}=A_hk_{p,h}$ denote a transformed key and $\bar q_{t,h}$ a post-RoPE query. At decoding,
the key must be restored by
$A_h^{-1}$ before applying its position-dependent RoPE rotation:
\begin{equation}
  \bar q_{t,h}^{\top}R_pk_{p,h}
  =\bar q_{t,h}^{\top}R_pA_h^{-1}u_{p,h}.
  \label{eq:diag-w-logit}
\end{equation}
When $A_h$ is diagonal, this restoration can be absorbed into the codebooks and adds no
runtime transform. In contrast, a dense $A_h^{-1}$ requires a $d\times d$ multiplication for every
cached key. It cannot be merged into the query projection because $R_p$ depends on the key
position. Table~\ref{tab:diag-w-cost} shows that this additional cost grows from $36\%$ of
attention time at 1k tokens to $103\%$ at 16k.

\begin{table}[h]
  \centering
  \small
  \begin{tabular}{rccc}
    \toprule
    Context & \multicolumn{2}{c}{Latency (ms/step)} & Overhead \\
    \cmidrule(lr){2-3}
    (tokens) & Attention & Dense Restoration & (\%) \\
    \midrule
     1\,024 & 1.55 & 0.55 & $+36\%$  \\
     4\,096 & 2.11 & 1.23 & $+58\%$  \\
    16\,384 & 4.14 & 4.27 & $+103\%$ \\
    \bottomrule
  \end{tabular}
  \caption{Cost of dense key restoration for Qwen3-4B-Thinking-2507 on one RTX 3090 at batch size
  1. Restoration is measured as a standalone BF16 GEMM over reconstructed keys, excluding VQ
  decoding; overhead is relative to attention latency.}
  \label{tab:diag-w-cost}
\end{table}

To test its effect on quantization quality, we fit two variants from the same calibration data using
either $\operatorname{diag}(\tilde{A}_h)^{1/2}$ or $\tilde{A}_h^{1/2}$, while holding all other
settings fixed. We then measure relative attention-logit error on the calibration corpus and
GSM8K, using
$\sum_{t,p}(\bar q_{t,h}^{\top}R_p\delta_{p,h})^2 /
\sum_{t,p}(\bar q_{t,h}^{\top}R_pk_{p,h})^2$.

\begin{table}[h]
  \centering
  \small
  \begin{tabular}{lcc}
    \toprule
    Key Transform & \multicolumn{2}{c}{Relative Attention-Logit Error} \\
    \cmidrule(lr){2-3}
    & Calibration Corpus & GSM8K \\
    \midrule
    $\operatorname{diag}(\tilde{A}_h)^{1/2}$ (TaSQ)
      & $0.00287$ & $\mathbf{0.00234}$ \\
    $\tilde{A}_h^{1/2}$
      & $\mathbf{0.00264}$ & $0.00933$ \\
    \bottomrule
  \end{tabular}
  \caption{Relative attention-logit error for Llama-3.1-8B-Instruct, averaged over $32$ layers.
  Both variants use the same calibration data and $K=1024$ codebooks.}
  \label{tab:diag-w-logit}
\end{table}

Dense weighting slightly lowers error on the calibration corpus but yields about $4\times$ higher
error on GSM8K. Its off-diagonal channel couplings do not transfer in this experiment.
The diagonal approximation therefore preserves efficient decoding while achieving lower error on
GSM8K.

\end{document}